\pdfoutput=1

\documentclass[11pt]{article}

\usepackage[final]{latex/acl}

\usepackage{times}
\usepackage{latexsym}
\usepackage[T1]{fontenc}
\usepackage[utf8]{inputenc}
\usepackage{microtype}
\usepackage{inconsolata}

\usepackage{hyperref}
\usepackage{url}

\usepackage{graphicx}
\usepackage{booktabs}

\usepackage{array}
\usepackage{amsmath}
\usepackage{amssymb}
\usepackage{mathtools}
\usepackage{multirow}
\usepackage{amsthm}
\usepackage{paralist}
\usepackage{subcaption}
\usepackage{xspace}

\usepackage{wela}

\newcommand{\todo}[1]{}
\newcommand{\signbetter}{\makebox[0pt]{\hspace{1.25ex}$^\star$}}
\newcommand{\signbetterboth}{\makebox[0pt]{\hspace{2.2ex}$^{\star\dag}$}}
\newcommand{\anomcitep}[1]{(Anonymous, 2023)\xspace}
\newcommand{\anomcitet}[1]{Anonymous (2023)\xspace}

\newcommand{\bleurt}{\textsc{Bleurt}\xspace}
\newcommand{\metricx}{MetricX\xspace}
\newcommand{\bleurtqe}{\metricx-{QE}\xspace}
\newcommand{\comet}{\textsc{Comet}\xspace}
\newcommand{\bleu}{\textsc{Bleu}\xspace}
\newcommand{\chrf}{ChrF\xspace}
\newcommand{\meteor}{Meteor\xspace}
\newcommand{\argmax}{\operatornamewithlimits{argmax}}

\newcommand{\qetag}[1]{\texttt{[}#1\texttt{]}}

\newcolumntype{H}{>{\setbox0=\hbox\bgroup}c<{\egroup}@{}}

\usepackage[capitalize,noabbrev]{cleveref}

\title{Quality-Aware Translation Models: Efficient Generation and Quality Estimation in a Single Model}

\usepackage{authblk}

\author[1,2,3]{{\bf Christian Tomani}\thanks{Work carried out while interning at Google.    Correspondence to: Christian Tomani (christian.tomani@tum.de), David Vilar (vilar@google.com).}}
\author[1]{\bf David Vilar}
\author[1]{\bf Markus Freitag}
\author[1]{\bf Colin Cherry}
\author[1]{\\\bf Subhajit Naskar}
\author[1]{\bf Mara Finkelstein}
\author[1]{\bf Xavier Garcia}
\author[2,3]{\bf Daniel Cremers}
\affil[1]{Google}
\affil[2]{Technical University of Munich}
\affil[3]{Munich Center for Machine Learning}

\begin{document}

\maketitle

\begin{abstract}

Maximum-a-posteriori~(MAP) decoding is the most widely used decoding strategy for neural machine translation~(NMT) models.
The underlying assumption is that model probability correlates well with human judgment, with better translations getting assigned a higher score by the model.
However, research has shown that this assumption does not always hold, and generation quality can be improved by decoding to optimize a utility function backed by a metric or quality-estimation signal, as is done by Minimum Bayes Risk~(MBR) or quality-aware decoding.
The main disadvantage of these approaches is that they require an additional model to calculate the utility function during decoding, significantly increasing the computational cost.
In this paper, we propose to make the NMT models themselves quality-aware by training them to estimate the quality of their own output.
Using this approach for MBR decoding we can drastically reduce the size of the candidate list, resulting in a speed-up of \emph{two-orders of magnitude}.
When applying our method to MAP decoding we obtain quality gains similar or even superior to quality reranking approaches, but with the efficiency of single pass decoding.

\end{abstract}

\section{Introduction}

Most state-of-the-art models for natural language processing tasks are probabilistic, with the most frequent parameterization being based on neural networks.
Once these models are trained, the prevailing decoding strategy for natural language generation is MAP decoding, i.e.\ select the hypothesis that maximizes the conditional probability given an input.
As an exact maximization is computationally intractable, typically beam search or greedy decoding are used to approximate the search for the best hypothesis.
Neural Machine translation is a prominent example of these types of models, where the system is trained to generate a sentence in a target language given a source sentence in another language.
Nonetheless, \citet{eikema-aziz-2020-map} have demonstrated that MAP decoding methods may be suboptimal due to the presence of misaligned probability distributions.
Moreover, 
NMT models often assign human translations lower probabilities than their own beam search outputs due to calibration issues~\citep{ott2018analyzing,freitag2020bleu}.

\citet{eikema-aziz-2020-map, eikema2022samplingbased} applied MBR decoding for NMT models as an alternative generation approach.
MBR decoding follows a self-consistency approach by sampling from the model distribution and giving preference to hypotheses that exhibit greater similarity to all other hypotheses.
In contrast to MAP decoding, MBR decoding's objective is not centered on generating the translation with the highest estimated model probability, instead it selects the translation that exhibits the highest quality based on a utility metric.
Subsequent research conducted by \citet{freitag2022high} showed that MBR decoding with \emph{neural} utility metrics leads to significant improvements over beam search decoding. 
However, MBR is computationally intensive, with a time complexity of $O(M^2)$ for a candidate list containing $M$ samples, ideally $M=100$ to $1\,000$ according to \citet{freitag2022high}.
Note than when using neural metrics, each ``computation step'' in the quadratic complexity is itself computationally expensive, requiring a forward pass through a large neural network.

As an alternative to MBR decoding, we can use a quality-aware decoding strategy, generating a list of candidate translations and reranking them using a neural quality estimation~(QE) metric that computes a quality score for the translation conditioned only on the source sentence. 
This method offers the advantage of being more efficient than MBR decoding, as its inference speed scales linearly with the number of candidate translations. 
A study conducted by \citet{fernandes-etal-2022-quality} showed that employing neural metrics for QE reranking exhibits comparable advantages to those seen with MBR decoding. 
However, this approach still demands the use of a separate, computationally expensive QE model to evaluate the quality of each candidate. 

In our work, we propose a novel method that moves quality awareness inside the model itself, enabling us to guide the generation process towards higher-quality translations, and eliminating the need for an external QE model during decoding.
Specifically, we investigate two key strategies: Quality-Aware Prompting, where we use quality prompts that explicitly encourage the generation of high-quality translations, and Quality-Aware Prediction, where we enable an NMT model to judge the quality of its own translations.
Both strategies add special quality tokens to each NMT training example. The strategies differ only in whether the token is included in the source or the target sentence. 

Our main scientific contributions are the introduction of quality-aware translation models, and their application for improved, more efficient decoding strategies.
We analyze two use cases:
\begin{itemize} 
    \item We propose a novel reranking approach that eliminates the necessity for external QE models during decoding while maintaining the same level of translation quality.
    We can achieve similar or even superior results with single pass decoding, eliminating the need of a costly reranking step.
    \item We show that pre-filtering the candidate list according to the model's quality prediction can dramatically boost decoding performance of MBR, by up to two orders of magnitude, while increasing translation quality.
\end{itemize}

\section{Related Work}

Machine Translation (MT) metrics can be divided into two high-level categories: reference-based and reference-free, also known as quality estimation (QE) metrics.
QE metrics compute a quality score for the MT output conditioned only on the source text.
Reference-based metrics, on the other hand, require a human-generated reference translation to compare the MT output with.
A multitude of reference-based metrics are available to evaluate the quality of translated content. Some metrics rely on lexical overlap, such as \bleu \citep{papineni-etal-2002-bleu}, \meteor \citep{banerjee-lavie-2005-meteor} or \chrf \citep{popovic-2015-chrf}. The WMT metrics task~\citep{freitag-etal-2022-results} demonstrated that the new generation of metrics -- neural fine-tuned metrics like \bleurt \citep{sellam2020bleurt} and \comet \citep{rei-etal-2020-comet} -- have significantly higher correlation with human judgement than traditional word overlap metrics. Consequently, we focus on neural fine-tuned metrics in this study.
Quality estimation for MT began as confidence estimation~\citep{blatz-etal-2004-confidence}, but has recently shifted to embrace close kinship with reference-based metrics, with recent neural examples including OpenKiwi~\citep{kepler-etal-2019-openkiwi}, TransQuest~\citep{ranasinghe-etal-2020-transquest}, and \comet-{QE}~\citep{rei-etal-2020-unbabels}.

A prominent example of a reference-based neural metric is \bleurt \citep{sellam2020bleurt}, and its extension to \metricx, which was the winning entry in the WMT22 metrics task \citep{freitag-etal-2022-results}. Building on this foundation, \citet{juraska-etal-2023-metricx} recently introduced \bleurtqe, which we use as our primary QE~metric in this work.

Reranking has a long history in translation, starting with \citet{shen-etal-2004-discriminative}, where a discriminative model is learned to rank a candidate list to maximize a reference-based metric, with recent examples including \citet{bhattacharyya-etal-2021-energy} and \citet{lee-etal-2021-discriminative}, who both train using BLEU. 
Our approach is closest to that of \citet{fernandes-etal-2022-quality}, who rerank using various translation quality-estimation metrics, as opposed to training a special-purpose discriminative reranker.
We differ from these works in that we do not need any external quality signal, which is instead provided by the NMT system itself.

While conventional MT research often relies on MAP decoding or generating k-best lists through beam search for MBR decoding, \citet{eikema-aziz-2020-map} proposed an approximation of MBR decoding via unbiased sampling.
Their method aims to address the limitations of MAP decoding \citep{eikema-aziz-2020-map, muller-sennrich-2021-understanding, eikema2022samplingbased} by demonstrating that samples drawn from the NMT model align more faithfully with training data statistics when compared to beam search.
\citet{freitag2022high} showed that using neural metrics instead of overlap metrics results in significant improvements in translation quality.
As a follow up, \citet{freitag2023epsilon} reported that the choice of sampling approach is important and epsilon sampling \citep{hewitt-etal-2022-truncation} is ideal for MBR decoding and reranking. \citet{cheng2023faster} introduced an orthogonal method to our proposed approach. They speed up MBR decoding by gradually increasing the number of samples used to estimate the utility while pruning hypotheses.

Our Quality-Aware Prompting approach extends a long line of methods where tagged training data has been used to control NMT output for different properties, including target language~\citep{johnson2016google},  formality level~\citep{yamagishi2016controlling}, politeness~\citep{sennrich-etal-2016-controlling}, domain~\citep{kobus-etal-2017-domain}, gender~\citep{vanmassenhove-etal-2018-getting}, syntactic structure~\citep{shu-etal-2019-generating}, complexity~\citep{agrawal-carpuat-2019-controlling} and reading level~\citep{marchisio-etal-2019-controlling}.
The  approaches closest to ours identify attributes related to translation quality, such as tagging back-translated examples to control away from synthetic data~\citep{caswell-etal-2019-tagged}, or tagging target-original examples to control toward natural-sounding output~\citep{freitag-etal-2022-natural}.
To the best of our knowledge, we are the first to use quality estimation to tag training data, allowing NMT to discriminate between different translation qualities, and allowing us to prompt the model to generate high quality translations.

\section{Preliminaries}

We are given a NMT model $P_{\Theta}(y|x)$ which serves to estimate the probability of a hypothesis segment $y \in \mathcal{Y}$, given a source segment $x$. 
Here, $\Theta$ denotes the learned parameters of the neural network and $\mathcal{Y}$ the set of all possible hypotheses.
There are two widely used approaches for generating the translations of a given sentence.

\paragraph{MAP decoding:}
This method involves searching for the most probable translation under $P_{\Theta}(y|x)$.
However, determining the hypothesis with the maximum probability is computationally intractable due to the expansive and combinatorially complex search space $\mathcal{Y}$.
Consequently, approximations like beam search \citep{graves2012sequence, sutskever2014sequence} are often employed.

\paragraph{Sampling:}
In many applications we want to generate diverse hypotheses, e.g.\ in generative tasks where creativity is desired.
In this case, instead of selecting the candidate with the highest probability (or an approximation thereof), we sample the output sentence following the probability distribution defined by the model.
For NMT, this approach is used for generating a list of candidate translations, e.g.\ for MBR decoding.
Specifically, epsilon sampling, as outlined by \citet{hewitt-etal-2022-truncation}, has emerged as the leading sampling technique for MBR. It was shown by \citet{freitag2023epsilon} to outperform other methods such as ancestral, top-k or nucleus sampling \citep{holtzman2019curious}.
Epsilon sampling prunes away any token with a probability lower than a given threshold $\varepsilon$, thereby guaranteeing that each token within a sample is allocated a fair probability mass. 
The likelihood of selecting token $y^{(\tau)}$ in the sampling process at time $\tau$ is governed by

\begin{equation}
    P'_{\Theta, \varepsilon} (y^{(\tau)}|x, y^{(1:\tau-1)}) \sim \begin{cases}
    p_\tau^{}{}^{\frac{1}{T}} & \text {if } p_\tau \geq \varepsilon \\
    0 & \text{otherwise}
    \end{cases},
\end{equation}
with
\begin{equation*}
p_\tau = P_{\Theta}(y^{(\tau)}|x, y^{(1:\tau-1)}).
\end{equation*}
$T$ denotes the sampling temperature.
Epsilon sampling proves to be a highly effective strategy for the selective removal of unreliable, low-probability tokens.

\subsection{External QE Reranking}

External QE-Reranking involves generating a candidate list of size $N$ through sampling and then reordering these samples, according to a quality estimation~(QE) model.
In our experiments, we employ \bleurtqe\footnote{\url{https://github.com/google-research/metricx}} \citep{juraska-etal-2023-metricx}, a modification of \metricx, to compute a quality score $q = f(x,y)$. Here, $f$ is parameterized by a transformer-based neural network and $x$ and $y$ denote the source segment and the translation, respectively. 

\subsection{Minimum Bayes Risk Decoding}
\label{sec:mbr}

In MBR decoding \citep{statistics1977basic,Berger_decision_theory_1985}, given a set of candidate hypotheses \(\mathcal{Y}\), the goal is to select the optimal hypothesis based on its expected utility concerning the distribution over human references within the space of all references \(\Omega\). This can be expressed mathematically as:

\begin{equation}
\label{eq:true_expected_utility}
h^{\rm best} 
=  \argmax_{y\in {\cal Y}} \sum_{r\in \Omega} u(y, r) P_{\rm human}(r|x),
\end{equation}
where $u(y, r)$ is a utility metric that is being used to gauge the quality of a candidate translation $y$ with respect to a reference translation $r$. 

Since \(P_{\text{human}}(r|x)\) remains unknown, we resort to sampling from the model instead, which relies on the assumption that the model provides a reliable approximation for the true underlying distribution over human translations.
Furthermore, the integration over the vast space of all possible references \(\Omega\) is computationally intractable. Therefore, MBR adopts a finite sample estimate by sampling a set of pseudo-references \(\mathcal{Y}_{\text{model}}\) from \(P_{\text{model}}(\cdot|x)\). This approximation can be expressed as:

\begin{equation}
\label{eq:approx_model_expected_utility}
h^{\rm MBR} = \argmax_{y \in {\cal Y}} \frac{1}{|{\cal Y_{\rm model}}|} \sum_{r\in \cal Y_{\rm model}} u(y, r),
\end{equation}
where \(\mathcal{Y} = \mathcal{Y}_{\text{model}}\), as the same set of model hypotheses serves both as the candidate list \(\mathcal{Y}\) as well as the pseudo-reference list \(\mathcal{Y}_{\text{model}}\).
The computational time complexity of MBR decoding is \(O(M^2)\) with $M$ the size of the candidate list.

Note that this quadratic expression refers to \emph{each sentence} to translate, i.e. for a corpus of size $S$, the total cost will be $O(S \cdot M^2)$.
Also there is a hidden (multiplicative) constant, namely the cost of the computation of the utility function.
For surface level metrics (e.g.\ \bleu, \chrf), this cost is negligible, but for neural metrics it involves computing the forward pass of a large neural network, 
therefore, any reduction in the number of metric computations has an important effect on the total running cost.
In this paper, we focus on using \bleurt as utility function during MBR decoding.

\section{Method}

\subsection{Quality-Aware Model}

In contrast to External QE Reranking, which uses a separate QE model for assessing the quality of translations, we propose a novel method that integrates quality awareness directly into the translation model, making an independent QE model unnecessary during decoding.
We present two approaches: in the first, we prompt the model to produce translations with a high QE score.
In the second, the model is designed to provide a quality score alongside the translation.
To achieve this, we initially assess the quality of samples within the training dataset, employing \bleurtqe.
In the training phase we train our NMT model simultaneously on source and target samples, as well as their associated quality scores.

\subsubsection{Assessing Quality}

We first prepare the training dataset by computing the translation quality of each training sample and labelling each sentence pair with the corresponding quality score.
Since the distribution over translation qualities is not necessarily uniform, we discretize the scores via equal mass binning into $B$ bins, which are then mapped to single tokens of the vocabulary of the translation model. This binning strategy leads to a balanced training set w.r.t.\ quality scores.
To achieve this, we consider the set of quality scores $Q$ from all samples in the training dataset, denoted as ${q_1, q_2, \ldots, q_N}$, in order to determine bin boundaries or cut-off points ${c_1, c_2, \ldots, c_{B+1}}$ such that $\forall 1 \leq i \leq B$

\begin{equation}
\left| \{x \; | \; q \in Q, \; c_i \leq q < c_{i+1} \} \right| \approx \frac{N}{B}.
\end{equation}

In this way each bin contains approximately the same number of samples.
This is to avoid sample imbalances per bin when training the model, which in preliminary experiments proved to be important as to not bias the model towards the most frequent label.
Next, each bin is assigned a bin identifier $b$, which is represented by a single token in the model vocabulary.
E.g.\ if we define 10 bins (the actual number we used in our experiments\footnote{For an exploration on other number of bins see Appendix~\ref{sec:number-of-bins}.}), we can just use the numbers between 0 and 9.
Lastly, the quality score is inserted into the data pipeline during training to associate the source and target pair with the respective bin identifier $b$.
To mark the token $b$ as a QE value, we employ a special string format by surrounding $b$ with square brackets:~\qetag{$b$}.
In the following we outline two distinct methods for integrating our quality score string into the model.

\subsubsection{Quality-Aware Prompting (QA Prompting)}
\label{sec:qa-prompting}

During the training process, we append the quality score string \qetag{$b$} to the source segment.
This enables the model to associate the discretized quality score $b$ with a translation example with the same level of quality.
As the quality token is attached to the \emph{input} to the system, it provides us with the capability to prompt for different quality levels during decoding.
I.e.\ at translation time we can append the token corresponding to the highest quality level to prompt the system to generate a sentence of the highest quality.

\subsubsection{Quality-Aware Prediction (QA Prediction)}
\label{sec:qa-prediction}

Instead of prompting the model explicitly for high-quality translations, an alternative approach is to design a model that jointly predicts a hypothesis and a quality score. 
This design allows us to leverage our translation model to also function as a QE model. 
To achieve this, we append the quality score string \qetag{$b$} to the target sentence during training, enabling the model to learn to predict the quality during inference.

If using a reranking approach these quality scores can be used to reorder samples within a candidate list.
However, due to our use of discretized bins, it is possible that the model predicts the same scores for multiple samples.
To address this, we also consider the log probabilities $z$ associated with the bin identifier tokens.
This additional information allows for a more precise reranking of samples in the candidate list.
Specifically, we sort samples with respect to $b$ as the primary sorting criterion and use the log probabilities $z$ as the secondary criterion.
Given a candidate list of size $M$, we sort the samples into ${y_1, y_2, \ldots, y_M}$ with corresponding discretized quality scores ${b_1, b_2, \ldots, b_M}$ and log probabilities ${z_1, z_2, \ldots, z_M}$ such that

\begin{equation}
\label{eq:sorting_reranking}
\begin{multlined}
\forall 1 \leq i < j \leq M: \\ (b_i > b_j) \lor \; (b_i = b_j \land z_i > z_j)
\end{multlined}
\end{equation}

With the sorted candidate list in hand, we can proceed by either selecting the top-ranked sample as our final translation or further processing the top-k samples in the context of MBR decoding.

\section{Experimental Setup}

\paragraph{Model:} 
Our model is a transformer consisting of 6 encoder and 6 decoder layers, 16 attention heads with a dimension of 128, a hidden dimension of 8192, and a model dimension of 1024, resulting in 551M parameters.
We employ a shared vocabulary of 32k tokens and impose a maximum sentence length of 128 tokens.
We utilize GELUs with gated activation functions.
The baseline system is trained on the entire available dataset.
All models are trained on TPUs (v3) until they reach convergence.
The \bleurtqe model used for quality estimation is a transformer based model with a total of 2B parameters, as described in \citet{juraska-etal-2023-metricx}.
To assess the applicability of our approach to LLMs, we also fine-tune and evaluate a quality-aware LLM, with results in Appendix~\ref{sec:appendix_llm}.

\paragraph{Data:} 
We choose two high-resource language pairs from the WMT 2022 shared task \citep{kocmi-etal-2022-findings}: English to German (en $\rightarrow$ de) and in the Appendix we additionally show results for English to Japanese (en $\rightarrow$ ja).
While we filter out sentences exceeding 128 tokens, we perform no further data filtering or preprocessing.
The training dataset for (en $\rightarrow$ de) comprises 295.8M samples, while the (en $\rightarrow$ ja) dataset consists of 33.9M samples. 
Our evaluation is based on the WMT 2022 general translation task test sets.

Given that we have available \bleurtqe scores for the whole training data, one natural question to ask is what would happen if we limit the training data to the only the best quality training sentence pairs.
\citet{peter-etal-2023-theres} showed that this is indeed a very effective way to reduce the training data size, while at the same time improving translation performance.
We also report experiments on this data condition, which represents a stronger baseline with which to compare our methods.
For these experiments we follow \citep{peter-etal-2023-theres} and keep only the top 50\% scoring sentence pairs.

\paragraph{Metrics: }
We use neural metrics for evaluation, with a focus on \comet \citep{rei-etal-2020-comet} (\comet~22 version).
We also report \metricx scores, but as \bleurtqe is based on it and our methods directly optimize this metric, there is the danger of overfitting for this particular metric \citep{amrhein-sennrich-2022-identifying,yan-etal-2023-bleurt}.
In addition, for selected experiments we conduct expert-based human evaluations using MQM \citep{freitag-etal-2021-experts}, a human evaluation scheme centered on marking errors present in the translations. Details can be found in Appendix~\ref{sec:mqm-evaluation}.
For completeness we also report \bleu scores in Appendix~\ref{sec:bleu-scores}, but we do not analyze them here, as it has been shown that they do not correlate well with human judgement of NMT systems \cite{freitag-etal-2022-results}.

\section{Results}

\subsection{Control Experiments}
\label{sec:control_experiments}

\subsubsection{Can NMT Models Learn to Estimate Quality?}

In a first experiment, we train a model using the proposed Quality-Aware Prediction method~(\S~\ref{sec:qa-prediction}).
We then use this model to force-decode the translations provided in the WMT23 QE shared task \cite{blain-etal-2023-findings} and, once the model has reached the end of the translation, we select the quality label that gets assigned the highest probability by the model.
As many different segments will be assigned the same discrete quality label, we also use model probability to break ties when calculating correlations (see Section~\ref{sec:qa-prediction}).
In this way we can check if our translation model can double as a QE model.
We evaluate this approach using the same metrics as in the shared task.
The results are shown in Table~\ref{tab:qe-correlations}, together with some other representative metrics.
The top block in Table~\ref{tab:qe-correlations} shows the top-scoring metrics in the shared task.\footnote{Note that evaluation metrics and QE metrics were evaluated simultaneously.}
As expected, we see that our model does not achieve the performance of separate dedicated models (and specifically not that of the \bleurtqe metric it is based on), but it still  outperforms the traditional \bleu and \chrf metrics, and these have access to the reference translations.

\begin{table}
    \centering
    \small
    \begin{tabular}{lrrr}
        \toprule
        & \multicolumn{2}{c}{Pearson $\rho$} & \\
        System &  \multicolumn{1}{c}{sys} & \multicolumn{1}{c}{seg} & \multicolumn{1}{c}{acc-t} \\
        \midrule
        XCOMET-Ensemble & 0.993 & 0.695 & 0.603 \\
        \metricx        & 0.977 & 0.585 & 0.602 \\
        \bleurtqe       & 0.969 & 0.626 & 0.596 \\
        \midrule
        QA Prediction (Ours) & 0.932 & 0.412 & 0.524 \\
        Model PPL            & 0.722 & 0.213 & 0.504 \\
        \midrule
        \bleu                & 0.916 & 0.192 & 0.520 \\
        \chrf                & 0.866 & 0.232 & 0.519 \\
        \bottomrule
    \end{tabular}
    \caption{Correlation of different QE metrics with human judgement on the WMT23 en $\rightarrow$ de task.}
    \label{tab:qe-correlations}
\end{table}

One could also consider using perplexity as an indicator of translation quality, thus eliminating the need of generating an explicit quality tag.
Table~\ref{tab:qe-correlations} shows that this value alone is not enough to differentiate the quality of the translations.

In summary, our quality-aware method can predict translation quality better than perplexity scores and older string-matching metrics, but not as well as dedicated neural models.
However, quality estimation is not the main goal of our work, rather a tool towards improving translation quality and efficiency, as we will show in Sections~\ref{sec:decoding_experiments} and~\ref{sec:mbr_experiments}.

\subsubsection{Can NMT Models Distinguish the Quality of Their Own Outputs?}

\begin{figure}
    \newlength{\figwidth}
    \setlength{\figwidth}{0.2\textwidth}
    \centering
    \includegraphics[width=0.47\textwidth]{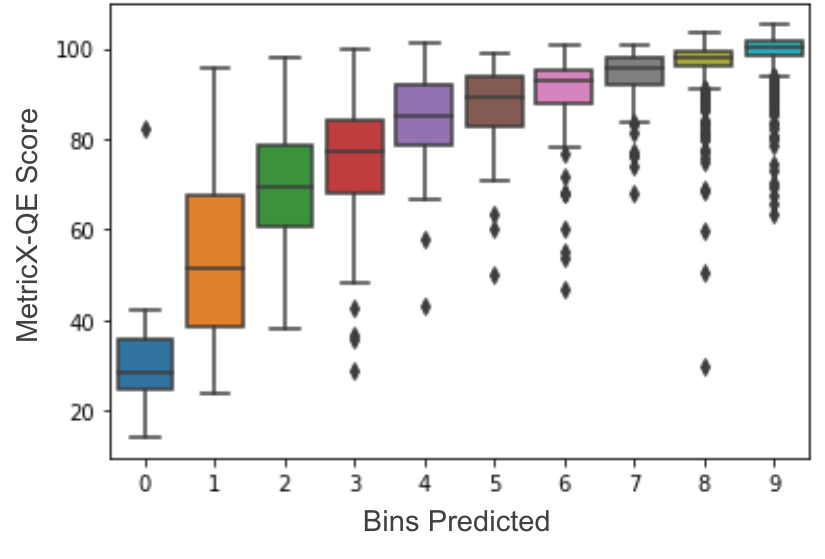}
	\caption{Alignment between predicted quality scores from the QA Prediction model and actual \bleurtqe scores of translations in the en $\rightarrow$ de test dataset. The boxplots show the distribution of actual scores across all samples assigned to each bin. The median ground truth quality score increases steadily in line with the predicted bins.}
\label{fig:score_corr_fig}
\end{figure}

Next we turn to the question of whether the system is able to judge the quality of \emph{its own} translations.
In order to test this, we translated our dev dataset with quality label predictions and then computed the actual \bleurtqe scores which the model is trained to predict.
We find that the predicted quality score bins are well aligned with these ``ground truth'' scores\footnote{``Ground truth'' in this context as these are the scores that the model was trained to predict. They are not necessarily ground truth for true quality measurement.}, as demonstrated in Figure~\ref{fig:score_corr_fig}, where we show the distribution of ground truth scores (non discretized) across the predicted bins. 
In particular we note that the system is able to accurately detect bad quality translations, which will become important for the translation use case.

\subsubsection{Can We Control Translation Quality?}

\begin{figure}
    \centering
    \includegraphics[width=\columnwidth]{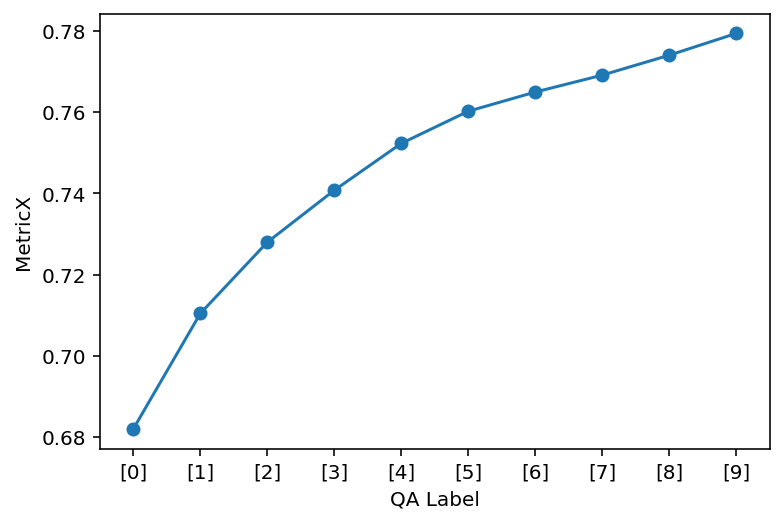}
    \caption{Translation quality dependent on the QA label used for prompting.
    Higher values in the label prompt the system to generate better translations.}
    \label{fig:qa_source_labels}
\end{figure}

As a last control experiment we train a QA Prompting model following Section~\ref{sec:qa-prompting} and generate translations using different quality labels.
I.e.\ we are asking the model to generate good translations (high scores in the quality labels) as well as \emph{bad} translations (low scores).
The results are shown in Figure~\ref{fig:qa_source_labels}.
It can be seen that the system is indeed able to adjust the quality of the translation output according to the given prompting.
Example translation outputs are shown in Appendix~\ref{sec:qa-prompting-examples}.

\subsection{Translation Performance}
\label{sec:decoding_experiments}

\begin{table*}[t]
\centering
\begin{tabular}{lccHcHc}
    \toprule
        Method & Data & \metricx & \bleurt & \comet & \bleu & MQM $\downarrow$\\
    \midrule
        Baseline & Full Corpus & 80.2 & 76.4 & 85.8 & 35.4 & 1.81 \\
        QA Prompting (Ours)& Full Corpus & \textbf{82.3} & 78.2 & \textbf{87.1}\signbetter & \textbf{36.6} & \textbf{1.43}\signbetter \\
        QA Prediction (Ours) & Full Corpus & 82.0 & \textbf{78.3} & 86.5\signbetter & 27.1 & 2.07 \\
    \midrule
        External QE-Reranking & Full Corpus & 83.3 & 78.4 & 86.9\signbetter & 28.1 & 1.50\signbetter \\
    \midrule
        Baseline & Filtered & 81.8 & 77.9 & 87.0 & 36.2 & -- \\
        QA Prompting (Ours) & Filtered & \textbf{82.6} & \textbf{78.5} & \textbf{87.3}\signbetter & \textbf{36.7} & -- \\
        QA Prediction (Ours) & Filtered & 82.5 & \textbf{78.5} & 86.7 & 28.1 & -- \\
    \midrule
        External QE-Reranking & Filtered & 83.7 & 78.6 & 86.9 & 27.8 & -- \\
    \bottomrule    
    \end{tabular}
\caption{
Comparison between quality-aware models and baseline models on the full and filtered training datasets.
The quality-aware methods outperform the baseline model and perform similar to reranking without requiring an additional \bleurtqe model during decoding.
$^\star$ denotes statistically significant (pairwise permutation test \cite{koehn-2004-statistical} with p=0.05) differences compared to the baseline with $p < 0.05$.
No significance is computed for \metricx due to the methods optimizing this metric directly.
}
\label{tab:quality_outputs}
\end{table*}

\begin{figure*}[t]
\centering
\begin{tabular}{cc}
\includegraphics[width=0.47\textwidth]{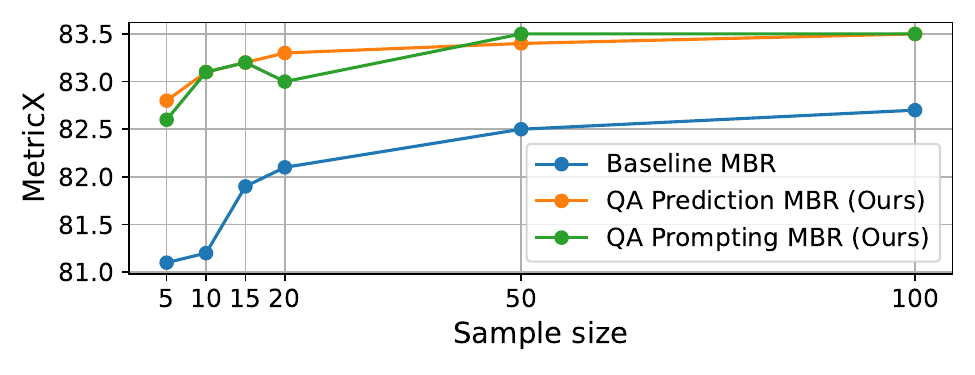} &
\includegraphics[width=0.47\textwidth]{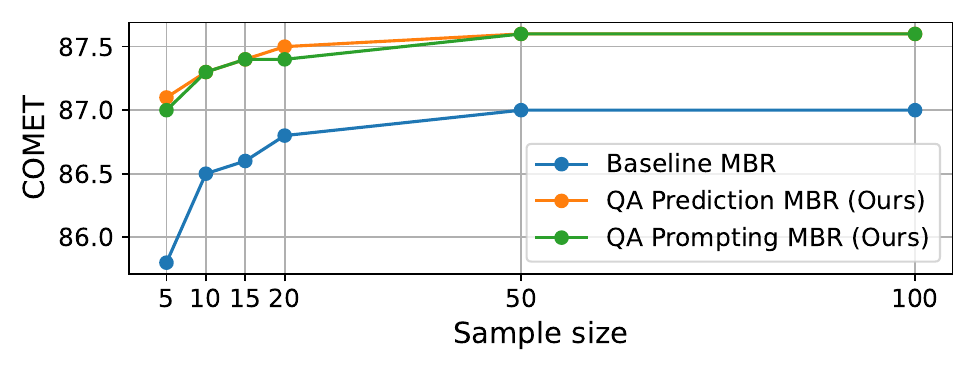} \\
{(a) \metricx - Full data} & {(b)  \comet - Full data}\\ 
\includegraphics[width=0.47\textwidth]{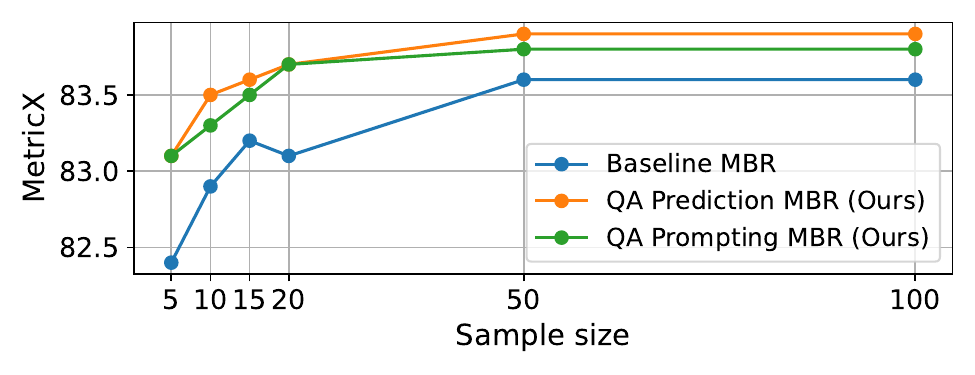} &
\includegraphics[width=0.47\textwidth]{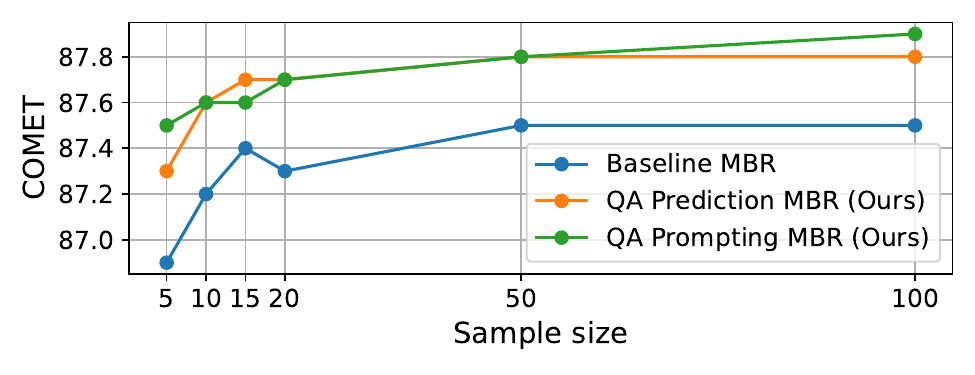} \\
{(c) \metricx - Filtered data} & {(d)  \comet - Filtered data}\\ 
\end{tabular}
\caption{Performance of quality-aware approaches (QA~Prediction and QA~Prompting) compared to baseline MBR decoding across various candidate list sizes. MBR decoding with quality-aware models consistently outperforms baseline MBR decoding across candidate list sizes. The quality-aware approaches can achieve the same level of performance as baseline approaches while reducing the required utility function computations by up to two orders of magnitude.}
\label{fig:mbr_line_plot}
\end{figure*}

Having confirmed that our model is indeed able to distinguish quality levels, we explore how to use this property to enhance the overall output quality of a NMT system.
For this experiment, we evaluate both our Quality-Aware Prediction model and Quality-Aware Prompting model.

For the Quality-Aware Prediction model, we extract $N$=$1024$ hypotheses from the model via epsilon sampling and retrieve the quality score string from each hypothesis.
Subsequently, we rank these hypotheses using Equation~\ref{eq:sorting_reranking} and select the highest-ranked sample as our final translation output.
In the case of Quality-Aware Prompting, we directly retrieve the final translation through MAP-decoding by appending the highest quality score string as a suffix to the source sentence, i.e.\ we ``ask the model'' to produce high quality outputs.
We compare these two approaches with an identical baseline NMT model that only differs by not using any quality score during training.
 
Table~\ref{tab:quality_outputs} shows that both quality-aware methods surpass the baseline model in terms of \metricx and \comet, with QA Prompting showing better results on both metrics.
As expected, all the methods achieve big improvements in \metricx, as it is closely related to the \bleurtqe metric that we are directly optimizing.
In fact, external QE-Reranking achieves the best \metricx score by a wide margin, however \comet puts QA Prompting on-par with QE-Reranking.
Even though an improvement of COMET of 0.2-0.4 seems small, \citet{lo-etal-2023-beyond} use human judgment to verify differences in COMET scores and conclude that even small differences in COMET scores can mean large quality improvements.

The human evaluation mostly confirms the trends shown by \comet.
The MQM scores can be interpreted as the average number of errors in translation, i.e.\ lower numbers are better.
The human evaluation shows that QA Prompting does indeed produce significantly better translations than the baseline systems, and it even outperforms the external QE-Reranking approach.
This is specially noteworthy given that QA Prompting is a single pass approach, with no additional cost over the MAP decoding baseline, whereas QE-Reranking rescores a 1024 candidate list with an external (expensive) QE model.
QA Prediction on the other hand does not outperform the baseline approach and in fact shows a degradation in performance.

When compared to the stronger baseline with training data prefiltered with \bleurtqe scores, we see that QA Prompting is still able to obtain a slight improvement over both the baseline and QE-Reranking, which is still statistically significant.
QA Prediction is not effective in this setting either.

\begin{table*}
\vskip 0.15in
\centering
\begin{tabular}{lllHcHcHc}
    \toprule
       Data & \#Candidates & Method & Data & \metricx & \bleurt & \comet & \bleu & MQM\\
    \midrule
    \multirow{8}{*}{\rotatebox{90}{Full Corpus}}
     && Baseline (w/o MBR) & Full & 80.2 & 76.4 & 85.8 & \textbf{35.4} & 1.81 \\
     && External QE-Reranking & Full & 83.3 & 78.4 & 86.9\signbetter & 28.1 & 1.50\signbetter \\
    \cmidrule{2-9}
    &\multirow{3}{*}{$M=50$}
        & MBR Baseline & Full & 82.5 & 80.3 & 86.8\signbetter & 30.8 & 1.41\signbetter \\
        && MBR QA Prompting (Ours) & Full & \textbf{83.5} & \textbf{81.1} & 87.4\signbetterboth & 31.9 & \textbf{1.36}\signbetterboth \\
        && MBR QA Prediction (Ours) & Full & 83.4 & 80.7 & \textbf{ 87.5}\signbetterboth & 31.1 & 1.45\signbetterboth \\
    \cmidrule{2-9}
    &\multirow{3}{*}{$M=5$}
        & MBR Baseline & Full & 81.1 & 78.2 & 85.7 & 29.6 & -- \\
        && MBR QA Prompting (Ours) & Full & 82.6 & \textbf{79.5} & 86.9\signbetter & 31.3 & -- \\
        && MBR QA Prediction (Ours) & Full & \textbf{82.8} & \textbf{79.5} & \textbf{87.0}\signbetter & 29.1 & -- \\
    \midrule
    \multirow{8}{*}{\rotatebox{90}{Filtered}}
        && Baseline (w/o MBR) & Filter & 81.8 & 77.9 & 87.0 & \textbf{36.2} & --\\
        && External QE-Reranking & Filter & 83.7 & 78.6 & 86.9 & 27.8 & --\\
    \cmidrule{2-9}
    &\multirow{3}{*}{$M=50$}
        & MBR Baseline & Filter & 83.6 & 81.0 & 87.4 & 31.1 & --\\
        && MBR QA Prompting (Ours) & Filter & 83.8 & \textbf{81.1} & \textbf{87.7}\signbetterboth & 33.0 & --\\
        && MBR QA Prediction (Ours) & Filter & \textbf{83.9} & 80.9 & \textbf{87.7}\signbetterboth & 32.0 & --\\
    \cmidrule{2-9}
    &\multirow{3}{*}{$M=5$}
        & MBR Baseline & Filter & 82.4 & 79.2 & 86.7 & \textbf{30.6} & -- \\
        && MBR QA Prompting (Ours) & Filter & \textbf{83.1} & \textbf{79.8} & \textbf{87.3}\signbetter & 32.4 & -- \\
        && MBR QA Prediction (Ours) & Filter & \textbf{83.1} & \textbf{79.8} & 87.2\signbetter & 30.2 & -- \\
    \bottomrule    
    \end{tabular}
\caption{
MBR results with quality-aware decoding approaches.
The symbol $^\star$ denotes statistically significant differences compared to the baseline with $p < 0.05$, $^\dag$ denotes statistically significant differences compared to the \emph{MBR baseline with $M=50$}.
No significance is computed for \metricx due to the methods optimizing this metric directly.
}
\label{tab:mbr_quality_both}
\vskip -0.1in
\end{table*}

\subsection{MBR decoding}
\label{sec:mbr_experiments}

Next we turn our attention to improving the performance of MBR decoding.
This depends heavily on the length of the candidate list $M$, and \citet{freitag2023epsilon} showed that a large candidate size of several hundred candidates is needed for achieving good translation performance.
However, this property makes MBR decoding computationally expensive as the utility function computations grow quadratically with the candidate list size~$M$, see Section~\ref{sec:mbr}.
Our investigation seeks to understand how improved candidate quality influences the performance and efficiency of MBR decoding.

For baseline MBR decoding we use epsilon sampling to generate a candidate list of~$M$ samples, as we do for the Quality-Aware Prompting approach.
For Quality-Aware Prediction, in line with the previous section, we employ our Quality-Aware Prediction approach to sample $N=1024$ hypotheses with quality score strings.
Then we rank all samples and select the top $M$ samples as our candidate list.
Subsequently, we apply MBR decoding to the gathered candidate lists.
We used \bleurt as utility function as a proxy for \metricx, due to the high computational cost of this last metric.

In Figure~\ref{fig:mbr_line_plot} we show the performance of our quality-aware approaches compared to baseline MBR decoding across various candidate list sizes~$M$.
Our proposed methods consistently outperform baseline MBR decoding in terms of \metricx and \comet scores, irrespective of the candidate list size.
Notably, our quality-aware approaches combined with MBR decoding require substantially fewer candidates to achieve equivalent performance to baseline MBR decoding.
For example, in models trained on the entire dataset, the Quality-Aware Prompting and Quality-Aware Prediction approaches obtain \comet scores of 86.9 and 87.0, respectively, with a candidate list size of 5 (needing just 20 utility function computations\footnote{The number of computations is $M \times (M-1)$ as a hypothesis is not evaluated against itself.} per sentence).
In contrast, baseline MBR decoding plateaus at 87 \comet starting at a candidate size of 50 (requiring 2450 utility function computations per sentence).
This translates to a 100-fold increase in computations for the baseline model to achieve a similar score.
We also note that the baseline model with a candidate list size of 1024 achieves a \metricx score of 82.9 and a \comet score of 87.0. 
This indicates that our approach, with a candidate list size of 50, outperforms a baseline model with even 1024 samples.
For our experiments on filtered data we observe a similar improvement in the quality-aware models when compared to baseline MBR decoding.

Table~\ref{tab:mbr_quality_both} shows the translation performance of the MBR systems, including human evaluation with MQM, with a candidate size $M = 50$.
A first observation is that baseline MBR decoding significantly outperforms the baseline system, coming close to the external QE reranking approach.
When we combine MBR with quality-aware models, we again obtain a significant improvement when compared the the MBR baseline, with the MQM score dropping from 1.81 to 1.36 for QA Prompting.
In this condition, the QA Prediction approach does perform satisfactorily, and there is no significant difference when compared to QA Prompting.

When reducing the candidate list size to 5, we can see that the translation performance drops only slightly (e.g.\ only 0.3 \comet for the Filtered QA Prompting approach), but the number of utility function computations is drastically reduced from 2450 to 20, two orders of magnitude.
This is not the case for baseline MBR, which actually performs worse than the non-MBR baseline with this reduced candidate size.

Note that each computation in MBR decoding involves evaluating a neural metric, an expensive operation.
Thus, reducing the absolute number of computation has a direct effect on running time of the full MBR pipeline.

\section{Conclusion}

This paper introduces a novel approach to enhance NMT by making the models quality-aware.
Our approach addresses the issue of misalignment between outputs generated via MAP decoding and human judgment.
We achieve this by training NMT models to assess the quality of their own translations, effectively circumventing the limitations of conventional decoding methods.
As a result this new approach yields improvements similar or superior to QE-reranking approaches, but with the efficiency of MAP-decoding, i.e.\ with single-pass decoding (for QA Prompting, the best performing method).
QE-reranking in contrast needs a sampling step followed by a reranking step using an external, computationally expensive model.

By leveraging the model's quality signal internally during MBR decoding, not only does translation quality further improve, but computational efficiency is also dramatically enhanced, reducing inference time by two orders of magnitude.
This improvement comes from a drastic reduction in the necessary size of the candidate list needed for producing good quality translations.

Overall this research opens up exciting possibilities for advancing the field of NMT, offering both improved translation quality and faster processing speeds without the need for additional, computationally intensive models.

\section*{Limitations and Risks}

Our work is currently limited to two language pairs. We leave it for future work to explore the applicability of the proposed approach in multilingual as well as low-resource settings. Furthermore, especially in low-resource languages where there is less training data, overfitting to the QE metric used for training could be a limitation. 

We acknowledge that, although we are using different metrics for optimizing our method (\bleurtqe) and evaluating it (\comet), both are neural metrics trained on the same data from the WMT evaluations.
There might be biases that should be taken into account when considering the method.
Nevertheless, neural metrics have proven to be the most reliable evaluation metrics for machine translation up to this date.

A potential risk of our method might be that training it is more resource-intensive than simple models, and thus might increase the quality difference with respect to low-resource languages, since they are unlikely to be allocated as many resources as the high-resource languages.

\bibliography{references}


\clearpage
\appendix

\section*{Appendices}

\section{Additional Results for Language Pair: English to Japanese}

Beyond the results highlighted in the main text, we present findings for an additional language pair, specifically, English to Japanese (\textbf{$en \rightarrow ja$}). Our experimental setup mirrors that of the English to German translation task, with the exception that we employ a \bleurt score threshold of 60 for training data filtering. Our results closely resemble those obtained for the English to German datasets.

Exploring the \textbf{$en \rightarrow ja$} scenario, we compare the performance of quality-aware models against baseline models using both the complete and filtered training datasets. Evaluation metrics include \metricx, \bleurt, \comet, and \bleu. Notably, the quality-aware methods achieve consistently better results than the baseline model, all without the need for an additional \bleurt model during decoding (Tab.~\ref{tab:quality_outputs_enja}).

\begin{figure*}
\centering
\begin{tabular}{cc}
\includegraphics[width=0.47\textwidth]{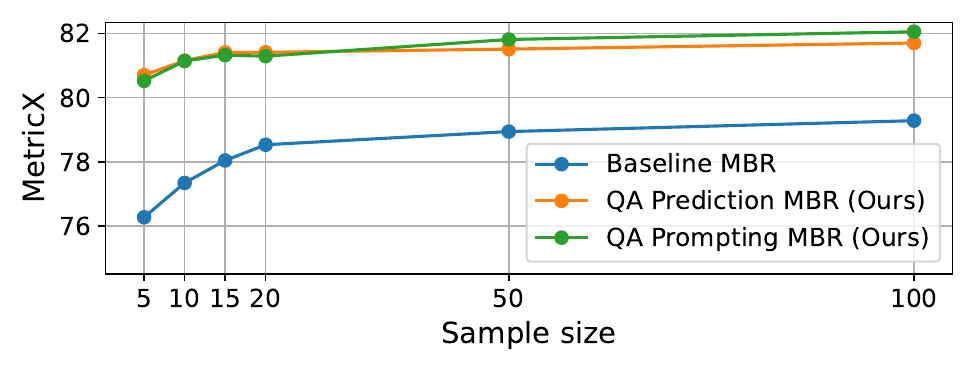} &
\includegraphics[width=0.47\textwidth]{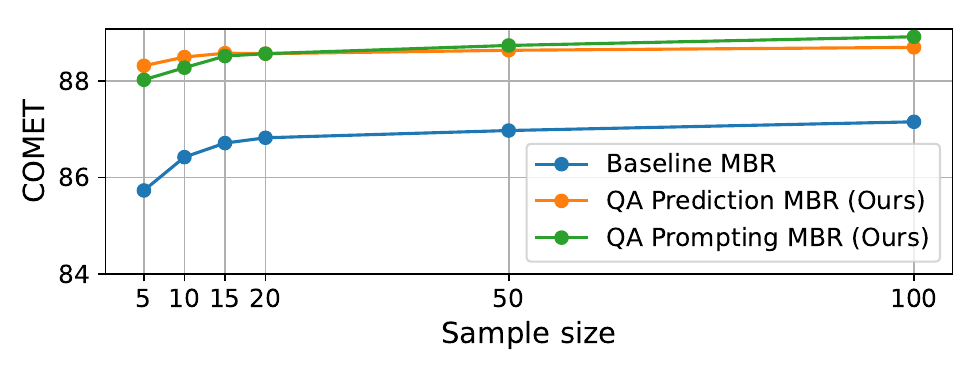} \\
{(a) \metricx - Full data} & {(b)  \comet - Full data}\\ 
\includegraphics[width=0.47\textwidth]{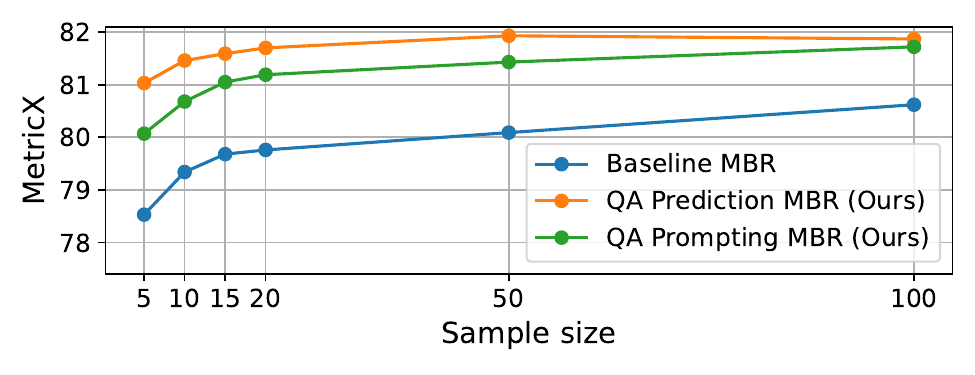} &
\includegraphics[width=0.47\textwidth]{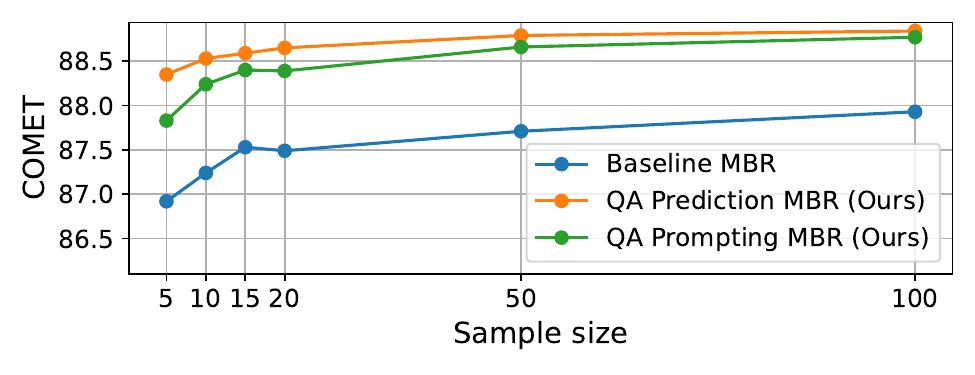} \\
{(a) \metricx - Filtered data} & {(b)  \comet - Filtered data}\\ 
\end{tabular}
\caption{Performance of quality-aware approaches (Quality-Aware Prediction and Quality-Aware Prompting) compared to baseline MBR decoding across various candidate list sizes for \textbf{$en \rightarrow ja$}. MBR decoding with quality-aware models consistently outperforms baseline MBR decoding across candidate list sizes. The quality-aware approaches can achieve the same level of performance as baseline approaches while reducing the required utility function computations by one to two orders of magnitude.}
\label{fig:mbr_line_plot_enja}
\end{figure*}

In Fig.~\ref{fig:mbr_line_plot_enja} we also examine the effectiveness of the proposed approaches in contrast to baseline MBR decoding across various candidate list sizes for \textbf{$en \rightarrow ja$}. Our findings demonstrate that MBR decoding with quality-aware models consistently surpasses baseline MBR decoding across different candidate list sizes. When trained on the entire dataset our methods achieve \metricx and \comet scores with only 5 samples that are clearly better than the baseline model regardless of the sample size. For this language pair we can dramatically increase the performance while at the same time requiring more than two orders of magnitude less computation time. When trained on the full dataset, our methods exhibit MetricX and \comet scores that outperform the baseline model for all candidate list sizes with a noticeable advantage, even when considering a limited number of samples. Importantly, this improved performance is accompanied by a similar reduction in computation time as for \textbf{$en \rightarrow de$}, with our approach requiring two orders of magnitude less computational resources for this language pair.

\begin{table*}[ht]
\vskip 0.15in
\centering
\begin{tabular}{lccHcH}
    \toprule
        Method & Data & \metricx & \bleurt & \comet & \bleu\\
    \midrule
        Baseline & Full Corpus & 76.3 & 65.5 & 85.7 & 19.5\\
        Quality-Aware Prompting (Ours)& Full Corpus & 80.3 & \textbf{67.2} & \textbf{87.8} & \textbf{23.6}\\
        Quality-Aware Prediction (Ours) & Full Corpus & \textbf{80.9}	& 66.6 & 87.7 & 18.9\\
    \midrule
        Baseline & Filtered & 77.8 & 64.9	& 86.2 & 21.4\\
        Quality-Aware Prompting (Ours) & Filtered & \textbf{80.0} & \textbf{66.9} & \textbf{87.7} & \textbf{23.1}\\
        Quality-Aware Prediction (Ours) & Filtered & 79.6 & 66.3 & 87.5 & 19.2\\
    \bottomrule    
    \end{tabular}
\vskip -0.1in
\caption{Comparison \textbf{$en \rightarrow ja$} between quality-aware models (Quality-Aware Prediction and Quality-Aware Prompting) and baseline models on the full and filtered training dataset evaluated on \metricx, \bleurt, \comet and \bleu. The quality-aware methods outperform the baseline model without the need of an additional \bleurt model during decoding.}
\label{tab:quality_outputs_enja}
\end{table*}

\section{BLEU scores}
\label{sec:bleu-scores}

For completeness we include in Tables~\ref{tab:quality_outputs-bleu} and~\ref{tab:mbr_quality_both-bleu} a copy of the main results of our paper, including \bleu scores.
However, as demonstrated in numerous previous work \citep[inter alia]{mathur-etal-2020-tangled,freitag-etal-2022-results,lo-etal-2023-beyond,freitag-EtAl:2023:WMT,kocmi2024navigating,zouhar2024quality} \bleu scores are not longer representative of translation quality with current systems, hence we do not base the analysis of our results on them.
Even more, Tables~\ref{tab:quality_outputs-bleu} and~\ref{tab:mbr_quality_both-bleu} provide further evidence of this fact, as \bleu scores show \emph{negative correlation} with our available human evaluations.

\newcommand{\badbleu}[1]{\textcolor{red}{#1}}

\begin{table*}[p]
\centering
\begin{tabular}{lccHccc}
    \toprule
        Method & Data & \metricx & \bleurt & \comet & \bleu & MQM $\downarrow$\\
    \midrule
        Baseline & Full Corpus & 80.2 & 76.4 & 85.8 & 35.4 & 1.81 \\
        QA Prompting (Ours)& Full Corpus & \textbf{82.3} & 78.2 & \textbf{87.1}\signbetter & \textbf{36.6} & \textbf{1.43}\signbetter \\
        QA Prediction (Ours) & Full Corpus & 82.0 & \textbf{78.3} & 86.5\signbetter & \badbleu{27.1} & \badbleu{2.07} \\
    \midrule
        External QE-Reranking & Full Corpus & 83.3 & 78.4 & 86.9\signbetter & \badbleu{28.1} & \badbleu{1.50}\signbetter \\
    \midrule
        Baseline & Filtered & 81.8 & 77.9 & 87.0 & 36.2 & -- \\
        QA Prompting (Ours) & Filtered & \textbf{82.6} & \textbf{78.5} & \textbf{87.3}\signbetter & \textbf{36.7} & -- \\
        QA Prediction (Ours) & Filtered & 82.5 & \textbf{78.5} & 86.7 & 28.1 & -- \\
    \midrule
        External QE-Reranking & Filtered & 83.7 & 78.6 & 86.9 & 27.8 & -- \\
    \bottomrule    
    \end{tabular}
\caption{
Copy of Table~\ref{tab:quality_outputs} including \bleu scores.
We highlight in red the cases where \bleu shows \emph{inverse correlation} with human judgements when comparing systems to the baseline.
}
\label{tab:quality_outputs-bleu}
\end{table*}

\begin{table*}[p]
\vskip 0.15in
\centering
\begin{tabular}{lllHcHccc}
    \toprule
       Data & \#Candidates & Method & Data & \metricx & \bleurt & \comet & \bleu & MQM\\
    \midrule
    \multirow{8}{*}{\rotatebox{90}{Full Corpus}}
     && Baseline (w/o MBR) & Full & 80.2 & 76.4 & 85.8 & \textbf{35.4} & 1.81 \\
     && External QE-Reranking & Full & 83.3 & 78.4 & 86.9\signbetter & \badbleu{28.1} & \badbleu{1.50}\signbetter \\
    \cmidrule{2-9}
    &\multirow{3}{*}{$M=50$}
        & MBR Baseline & Full & 82.5 & 80.3 & 86.8\signbetter & \badbleu{30.8} & \badbleu{1.41}\signbetter \\
        && MBR QA Prompting (Ours) & Full & \textbf{83.5} & \textbf{81.1} & 87.4\signbetterboth & \badbleu{31.9} & \badbleu{\textbf{1.36}}\signbetterboth \\
        && MBR QA Prediction (Ours) & Full & 83.4 & 80.7 & \textbf{ 87.5}\signbetterboth & \badbleu{31.1} & \badbleu{1.45}\signbetterboth \\
    \cmidrule{2-9}
    &\multirow{3}{*}{$M=5$}
        & MBR Baseline & Full & 81.1 & 78.2 & 85.7 & 29.6 & -- \\
        && MBR QA Prompting (Ours) & Full & 82.6 & \textbf{79.5} & 86.9\signbetter & 31.3 & -- \\
        && MBR QA Prediction (Ours) & Full & \textbf{82.8} & \textbf{79.5} & \textbf{87.0}\signbetter & 29.1 & -- \\
    \midrule
    \multirow{8}{*}{\rotatebox{90}{Filtered}}
        && Baseline (w/o MBR) & Filter & 81.8 & 77.9 & 87.0 & \textbf{36.2} & --\\
        && External QE-Reranking & Filter & 83.7 & 78.6 & 86.9 & 27.8 & --\\
    \cmidrule{2-9}
    &\multirow{3}{*}{$M=50$}
        & MBR Baseline & Filter & 83.6 & 81.0 & 87.4 & 31.1 & --\\
        && MBR QA Prompting (Ours) & Filter & 83.8 & \textbf{81.1} & \textbf{87.7}\signbetterboth & 33.0 & --\\
        && MBR QA Prediction (Ours) & Filter & \textbf{83.9} & 80.9 & \textbf{87.7}\signbetterboth & 32.0 & --\\
    \cmidrule{2-9}
    &\multirow{3}{*}{$M=5$}
        & MBR Baseline & Filter & 82.4 & 79.2 & 86.7 & \textbf{30.6} & -- \\
        && MBR QA Prompting (Ours) & Filter & \textbf{83.1} & \textbf{79.8} & \textbf{87.3}\signbetter & 32.4 & -- \\
        && MBR QA Prediction (Ours) & Filter & \textbf{83.1} & \textbf{79.8} & 87.2\signbetter & 30.2 & -- \\
    \bottomrule    
    \end{tabular}
\caption{
Copy of Table~\ref{tab:mbr_quality_both} including \bleu scores.
We highlight in red the cases where \bleu shows \emph{inverse correlation} with human judgements when comparing systems to the baseline.
}
\label{tab:mbr_quality_both-bleu}
\vskip -0.1in
\end{table*}

\section{Quality-Aware LLMs}

In this section we explore the potential of enhancing LLMs with quality awareness through our proposed method.
Specifically, we employ our most efficient approach, Quality-Aware Prompting, to finetune PaLM-2 Bison \cite{anil2023palm}. 
To this end, we finetune the pretrained model for 10k steps with the same data as in the main text (en-de) using our QA Prompting approach.
As baseline we also finetune the LLM with identical configurations on the same data, but without incorporating any quality signal.
In Table~\ref{tab:llm_finetuning}, we observe that QA Prompting 
substantially outperforms standard finetuning by 1.3 \bleurt points.
Our results suggest that substantial performance improvements can be achieved with minimal data and steps.
This opens the door to the possibility of utilizing more costly QE methods or even human evaluations in the future to curate finetuning datasets and align models directly with human preferences.

\label{sec:appendix_llm}
\begin{table}[ht]
\vskip 0.15in
\centering
\begin{tabular}{lcH}
    \toprule
        Method (LLM) & \bleurt & \bleu \\
    \midrule
        Baseline & 77.9 & 36.6\\
        QA Prompting (Ours) & \textbf{79.2} & \textbf{37.3}\\
    \bottomrule    
    \end{tabular}
\caption{Quality-Aware LLMs: The performance of LLMs can be enhanced via finetuning a pretrained LLM with Quality-Aware Prompting compared to standard finetuning.}
\label{tab:llm_finetuning}
\vskip -0.1in
\end{table}

\section{Sensitivity Analysis of Number of Bins}
\label{sec:number-of-bins}

\begin{figure*}
\centering
\begin{tabular}{ccc}
\includegraphics[width=0.31\textwidth]{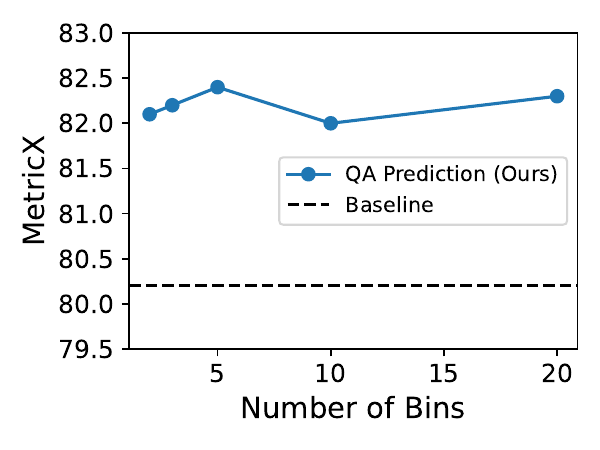} &
\includegraphics[width=0.31\textwidth]{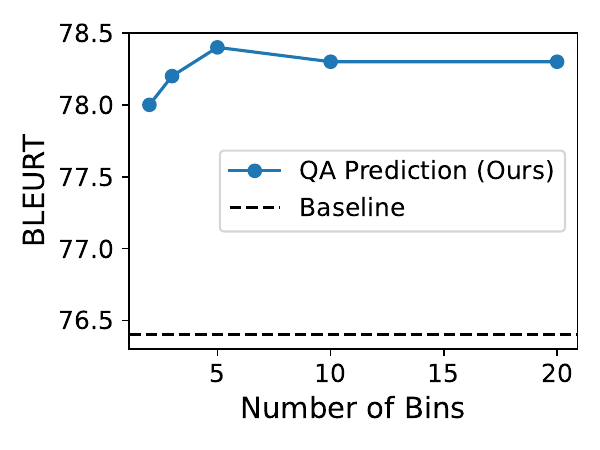} &
\includegraphics[width=0.31\textwidth]{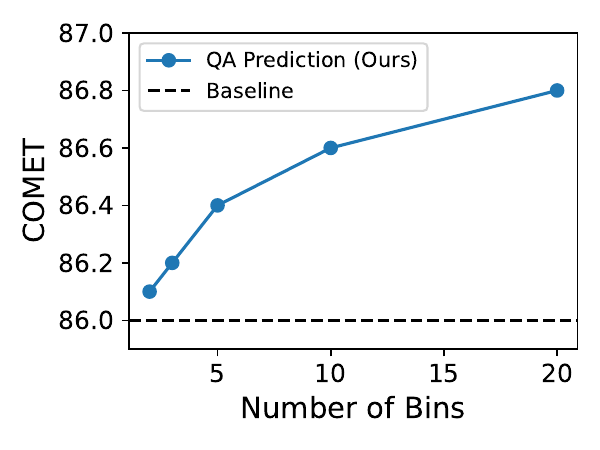} \\
\end{tabular}
\caption{Sensitivity concerning performance of the Quality-Aware Prediction approach w.r.t. the number of quality score bins. Increasing the number of quality score bins yields generally improvements on our quality metrics, specifically in the range of 2 to 5 bins.}
\label{fig:sensitivity_no_bins_line_plot}
\end{figure*}

In this section, we examine the selection of the number of bins for discretizing the quality score. To do this, we employ our Quality-Aware Prediction approach and train five models with varying numbers of bins, specifically 2, 3, 5, 10, and 20. We then evaluate their performance on \metricx, \bleurt, \comet, and \bleu. 
Figure~\ref{fig:sensitivity_no_bins_line_plot} illustrates that increasing the number of bins yields improvements on our quality metrics, in particular in the range of 2 to 5 bins. This aligns with our overarching concept of instilling quality awareness in the model, as a higher number of bins allows for a finer distinction between quality levels within the model, which is evident in our findings.

\section{Influence of Bin Identifier Choice}

Tokens in a language model are mapped to an embedding, representing a specific meaning or relation to other tokens within the embedding space. In this context, we aim to explore whether mixing token meanings from the primary translation task with the scoring task has a detrimental effect on either one of them. To investigate this, we assess the translation quality of the baseline model without the inclusion of quality score strings, as well as our proposed Quality-Aware Prediction approach.

Firstly, we evaluate the translation quality between the baseline and our model. We find that our model performs equivalently to the baseline when treating it as a basic translation model and disregarding the quality score strings appended to our model's hypotheses. This suggests that the model successfully disentangles a token's actual meaning in the text from its role as a quality score bin identifier.

Furthermore, we investigate whether the use of tokens that appear frequently in the training data corpus, such as numbers and letters, as opposed to tokens that are the least likely to appear in the training data, has an adverse impact on quality scoring. To investigate this, we train our model with 5 bins but employ different bin identifiers, including numbers $[0, 1, 2, 3, 4]$, letters $[a, b, c, d, e]$, and the 5 least frequently occurring tokens from the vocabulary. Tab.~\ref{tab:sensitivity_bin_ids} shows that varying the choice of bin identifiers demonstrates a high degree of robustness. 

We further investigate possible cases where the model does not produce a quality score at all. However, we find that the model is able to learn to predict a quality score as early as after 500 steps. We only observe a few corner cases where the model does not predict a quality score when we use epsilon sampling and sample 1000 sentences. In the case where no quality score is produced we assign the lowest bin identifier to a sample.

\begin{table*}[ht]
\vskip 0.15in
\centering
\begin{tabular}{lcHcH}
    \toprule
        Bin identifiers & \metricx & \bleurt & \comet & \bleu \\
    \midrule
        Baseline & 80.2 & 76.4 & 86.0 & 35.4 \\
    \midrule
        Numbers & 82.4 & 78.4 & 86.4 & 25.3 \\
        Letters & 82.5 & 78.5 & 86.6 & 25.5 \\
        Least likely tokens & 82.6 & 78.5 & 86.6 & 25.9\\
    \bottomrule    
    \end{tabular}
\caption{Sensitivity of our Quality-Aware Prediction approach with respect to bin identifiers on the entire training dataset. For all experiments we set the number of bins to 5 and vary the choice of bin identifiers. Our approach is largely insensitive to bin identifiers.}
\label{tab:sensitivity_bin_ids}
\vskip -0.1in
\end{table*}

\section{Example Translations for QA Prompting}
\label{sec:qa-prompting-examples}

In Table~\ref{tab:qa_promping_examples} we provide different translation examples when varying the label for quality prompting (see Section~\ref{sec:qa-prompting}).
It can be seen, that the prompting has a critical influence in the quality of the translation output.

\begin{table*}
    \centering
    
    \newcommand{\annotate}[1]{\textcolor{blue}{\wela \small #1}}
    
    \begin{tabular}{llp{10cm}}
    \toprule
    \multicolumn{2}{l}{Source} & Needless to say, it was chaos. \\
    \midrule
    \multirow{3}{*}{QA Label} &
    \qetag{0} &  Es war ein sehr schöner Tag. \annotate{Translation: "It was a very nice day."} \\
    & \qetag{5} &  Es war ein Chaos. \annotate{Missing translation for "Needless to say".} \\
    & \qetag{9} &  Unnötig zu sagen, es war Chaos. \annotate{Nearly perfect translation.} \\
    \midrule
    \multicolumn{2}{l}{Reference} & Es war natürlich ein Chaos. \\
    \bottomrule
    \end{tabular}
    
    \vspace*{1em}
    
    \begin{tabular}{llp{10cm}}
    \toprule
    \multicolumn{2}{l}{Source} & The total number of vaccines injected in the country reached 109,990,742 doses. \\
    \midrule
    \multirow{6}{*}{QA Label} &
    \qetag{0} &  Die Gesamtzahl der injizierten Impfstoffe in der Türkei erreichte 109.990.742 Dosen. \annotate{"in der Türkei" means "in Turkey"} \\
    & \qetag{5} &  Die Gesamtzahl der injizierten Impfstoffe erreichte 109.990.742 Dosen. \annotate{Missing the information about the country.} \\
    & \qetag{9} &  Die Gesamtzahl der injizierten Impfstoffe im Land erreichte 109.990.742 Dosen. \annotate{Perfect translation.} \\
    \midrule
    \multicolumn{2}{l}{Reference} & Die Gesamtzahl der im Land verabreichten Impfungen beläuft sich auf 109.990.742 Dosen. \\
    \bottomrule
    \end{tabular}
    
    \vspace*{1em}
    
    \begin{tabular}{llp{10cm}}
    \toprule
    \multicolumn{2}{l}{Source} & Long story short, she comes back to the checkout, but goes to another line. \\
    \midrule
    \multirow{7}{*}{QA Label} & 
    \qetag{0} & Kurz gesagt, sie kommt zurück zur Kasse, aber geht zu einer anderen Zeile. \annotate{"Zeile" is an incorrect word in this context.} \\
    & \qetag{5} & Kurz gesagt, sie kommt zurück zur Kasse, geht aber zu einer anderen Linie. \annotate{"Linie" is better, but still not natural.} \\
    & \qetag{9} & Lange Rede kurzer Sinn, sie kommt zurück zur Kasse, geht aber zu einer anderen Schlange. \annotate{"Schlange" is the correct word. Also nice use of the idiom "Lange Rede kurzer Sinn".} \\
    \midrule
    \multicolumn{2}{l}{Reference} & Um es kurz zu machen, sie kam zurück zur Kasse, stellte sich aber bei einer anderen Schlange an. \\
    \bottomrule
    \end{tabular}
    \caption{Annotated example translations for different quality labels used for prompting.}
    \label{tab:qa_promping_examples}
\end{table*}

\section{Combining Quality-Aware Prompting and Quality-Aware Prediction}

Throughout our experiments we frame Quality-Aware Prompting and Quality-Aware Prediction as two separate approaches. One might wonder whether both approaches are orthogonal to each other and might benefit each other when combined. To this end we add the quality score string to the source and the target sentence. 
To avoid that the model just learns copying the quality score string from the input to the output, we choose a multiple of the bin number from the prompting approach for the prediction approach. This way we make sure that the model is required to first of all learn to provide a high quality translation when prompted for it and learns to fine grained distinguish the quality in the quaility prediction tasks in the output. 
However, we observe that the combination of both approaches results in inferior performance compared to each approach individually, regardless of whether the full or filtered dataset is employed for training (Tab.~\ref{tab:abl_comb_prompt_pred}). We hypothesize that this may be attributed to the model becoming excessively fixated on predicting the score in the output based on the input score, potentially leading to overfitting, where the prediction score becomes overly conditioned on the prompting score.

\begin{table*}[ht]
\vskip 0.15in
\centering
\caption{Performance of combining Quality-Aware Prompting and Quality Prediciton approaches on both full and filtered datasets. Two combined models are trained for each dataset: one with 10 and the other with 20 prediction quality score bins, while using 5 quality score bins for prompting. Combining both approaches yields no improvements across metrics.}
\label{tab:abl_comb_prompt_pred}
\begin{tabular}{lccHcH}
    \toprule
        Method & Data & \metricx & \bleurt & \comet & \bleu \\
    \midrule
        Quality-Aware Prompting & Full & 82.3 & 78.2 & 87.1 & 36.6 \\
        Quality-Aware Prediction & Full & 82.0 & 78.3 & 86.6 & 27.1 \\
        Combo: ProBins=5 and PreBins=10 & Full & 82.1 & 78.4 & 86.7 & 27.4 \\
        Combo: ProBins=5 and PreBins=20 & Full & 82.1 & 78.1 & 86.7 & 28.8 \\
    \midrule
        Quality-Aware Prompting & Filter & 82.6 & 78.5 & 87.3 & 36.7 \\
        Quality-Aware Prediction & Filter & 82.5 & 78.5 & 86.9 & 28.1 \\
        Combo: ProBins=5 and PreBins=10 & Filter & 81.9 & 78.0 & 86.6 & 28.5 \\
        Combo: ProBins=5 and PreBins=20 & Filter & 82.1 & 78.2 & 86.8 & 29.4 \\
    \bottomrule    
    \end{tabular}
\vskip -0.1in
\end{table*}

\section{Details of Quality-Aware Approaches}

\begin{figure*}[ht]
    \centering
    \begin{subfigure}{0.75\textwidth}
    \includegraphics[width=\textwidth]{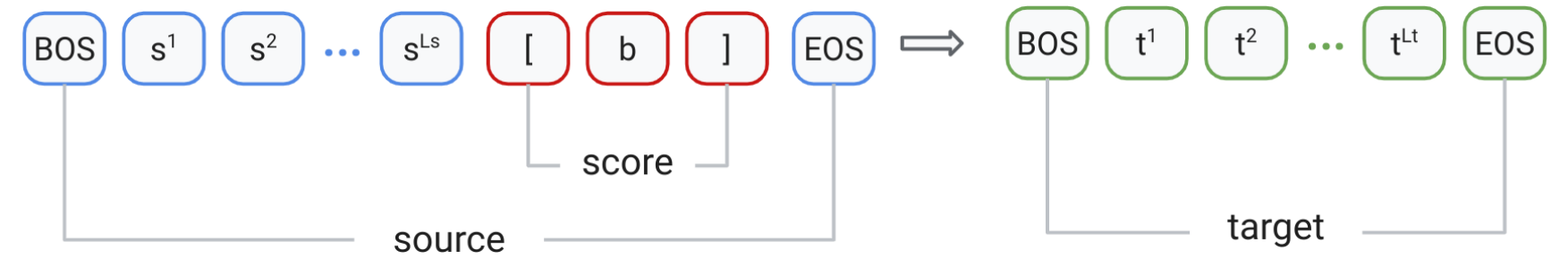}
    \subcaption{Quality-Aware Prompting}
    \label{fig:visualization_prompting}
    \end{subfigure}
    
    \begin{subfigure}{0.75\textwidth}
    \includegraphics[width=\textwidth]{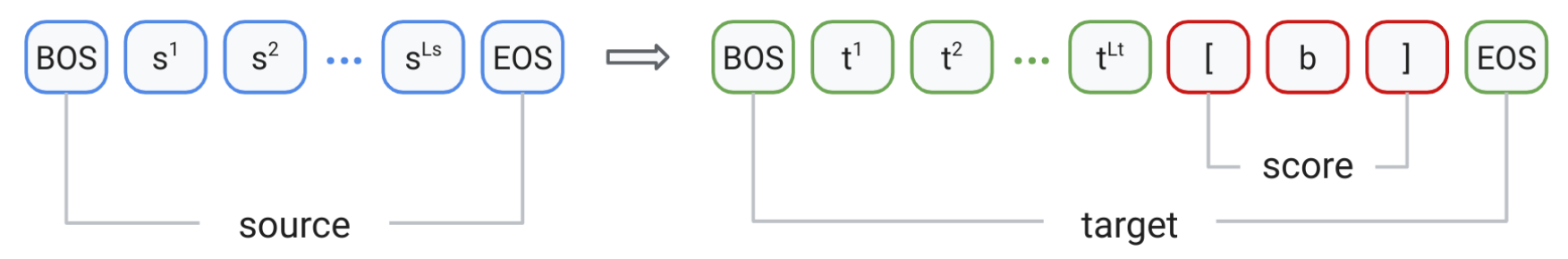}
    \subcaption{Quality-Aware Prediction}
    \label{fig:visualization_prediction}
    \end{subfigure}
	\caption{Visualization of the Quality-Aware Prompting and Quality-Aware Prediction approaches.}
\label{fig:visualization_quality_aware_models}
\end{figure*}

In the context of quality-aware translation, two approaches are explored. In Quality-Aware Prompting (QA Prompting), a quality score is appended to the source segment during training, allowing the model to associate quality score strings with examples of translation exhibiting the same level of quality
(Figure~\ref{fig:visualization_prompting}). Since this quality token appears in the input, it allows for direct prompting of high-quality translations during decoding. Conversely, Quality-Aware Prediction (QA Prediction) involves training a model that predicts both a hypothesis and a quality score concurrently. This approach transforms the translation model into a Quality Estimation (QE) model by appending the quality score string to the target sentence during training, enabling the model to predict quality during inference (Figure~\ref{fig:visualization_prediction}).

\section{MQM Human Evaluations}
\label{sec:mqm-evaluation}

For the MQM evaluation we use the same annotator guidelines as specified in \citep{freitag-etal-2021-experts}. The annotators were compensated fairly and did
not have to disclose any personal information during the annotation process. All of the test sets used in this study are publicly available.


\end{document}